\documentclass[twoside,11pt]{article}

\usepackage[preprint]{jmlr2e}

\usepackage[utf8]{inputenc}
\usepackage[T1]{fontenc}
\usepackage{times}          
\usepackage{microtype}
\usepackage{booktabs}
\usepackage{amsmath}
\usepackage{url}
\usepackage{float}
\usepackage{placeins}
\usepackage{needspace}
\usepackage[bottom]{footmisc}

\newcommand{\pgbrk}{\hfill\vadjust{\pagebreak[4]}\break}
\firstpageno{1}

\begin{document}

\title{When Does Self-Supervised Pretraining Help Tabular Models?\\
A Study of Label Scarcity and Missing Data}

\author{\name Sahand Mazrouei \email sahand.mazrue@khu.ac.ir \\
        \addr Faculty of Mathematics and Computer Science, Kharazmi University}

\maketitle

\begin{abstract}%
In this work, we conduct an empirical study on self-supervised learning (SSL) for tabular data. Our experiments examine limited label regimes (that is, 1\%, 5\%, 10\%, and 20\% of the labels) and highlight limitations and opportunities in current tabular SSL. Contrary to what is expected, our mask-and-recover approach shows its largest improvements for datasets that have little to no recorded numeric missingness, whereas it more often leads to degradation on datasets that are characterized by native numeric missingness (post hoc split). Random Forest achieves an average AUC of 0.9015 across the 14 tasks for the 10\% labels setting, whereas SSL shows a performance of $\sim$0.8954 (RF: 1 draw; SSL: 3-draw mean). Moreover, the gain of SSL compared to training from scratch exhibits a high variance across tasks and is not significant (1 run; Wilcoxon $p = 0.626$, Holm-adjusted $p = 1.000$, for both the 5\% and 10\% settings). On the other hand, we observe that SSL-pretrained models achieve a higher average AUC when MNAR and MCAR missingness were injected at test-time. In comparison to models trained from scratch, our SSL-pretrained models show an improvement of $+0.0245$ (MCAR) and $+0.0418$ AUC (MNAR) under test-time missingness (single seed), with 11 (MCAR; plus one tie) and 8 (MNAR) positive results across the 14 tasks. However, after Holm-Bonferroni correction neither improvement is significant (Holm-adjusted $p = 0.118$ for MCAR and $0.518$ for MNAR). Finally, when comparing mask-and-recover with three other approaches, namely, simplified VIME, SCARF and SubTab implementations using an identical encoder, we observe no significant difference in performance (Holm-adjusted $p = 0.459$ for VIME, $1.000$ for SCARF, $1.000$ for SubTab).
\end{abstract}


\section{Introduction}
\label{sec:intro}

Tabular data is a common way to represent heterogeneous data in applications ranging from financial forecasting to healthcare.
Classical, tree-based models such as Random Forest or Gradient Boosted Trees are commonly used for prediction on such data, often
outperforming neural networks on tabular tasks despite the success of deep learning in other fields~\citep{grinsztajn2022tree}. This gap between classical and neural models may become larger in realistic scenarios with limited labels and missing features: Labels can be hard to collect due to slow or costly annotation, and real-world data
 regularly features missingness due to irregular reporting, technical problems or privacy concerns.

Self-supervised learning offers a promising approach to addressing these challenges~\citep{vime,scarf,subtab}. Instead of a task-specific objective, SSL
optimizes a proxy-task, usually of the ``mask, corrupt, and recover'' family, leveraging large quantities of unlabeled data.
By training a self-supervised objective on abundant, unlabeled features, deep learning models can learn an encoder that captures the structure
of the underlying data manifold which can be adapted for prediction given few labeled data. Recent works have investigated the design of
self-supervised objectives for tabular data, but their performance in conditions of native missingness and test-time missingness shifts needs further study. How these objectives
interact with missingness that is present in the data itself, and how robust the learned representations are to missingness shifts at test-time remain open questions.
We seek to address these questions through a systematic empirical study of SSL for tabular data in a more realistic setting: We build a
suite of classification tasks from OpenML to evaluate SSL in scenarios of limited labels using a fixed 1\% ,5\% , 10\%,  20\% budget and stress-test robustness
by injecting missingness at test-time. We compare a self-supervised learner to common baselines, as well as its underlying neural architecture trained end-to-end
on the prediction objective. Looking past overall performance rankings, this investigation evaluates how SSL actually operates across varied datasets. Interestingly, although a preliminary analysis with only one random seed suggested an effect of ablating consistency loss (Section~\ref{sec:ablation}), re-doing this analysis with three separate pretraining seeds (the main label scarcity results use a single shared pretraining run), shows that this effect cannot be distinguished from zero, with mean $\Delta$AUC $= +0.0031$ and $p = 0.625$. In general, we did not observe a reliable effect of any of the ablations in our study, including consistency loss, group masking, and mask-visibility, that generalizes across the four native-missingness datasets. These findings highlight the risk of conducting studies with a single seed, particularly in the current low-sample regime of tabular SSL benchmarking.

We find that using 10\% of the labels, SSL-pretraining performs comparably to a strong, commonly used supervised learning algorithm, Random Forest, with mean AUC-ROCs within 0.0061 of RF (neural: mean of three draws; RF: one draw; untested), but exhibits high inter-task variance in its advantage over training from scratch, and that this advantage is not statistically significant.

We find that SSL shows its largest benefits for downstream classification on datasets with numerically cleaner data as opposed to datasets with native missingness (post hoc, descriptive), where the benefits of SSL are highly variable between datasets. However, when faced with distributional shifts at inference, SSL-pretrained models achieve a higher average AUC compared to models trained from scratch. Specifically, when facing added Missing Completely At Random (MCAR) missingness at a high amount of 30\% as well as distributional shift via structured Missing Not At Random (MNAR) changes, SSL-pretrained models have higher AUC on corrupted data than models trained from scratch (including their clean-data advantage of $+0.0086$ at 10\% labels; single fine-tuning seed), although this difference is not statistically significant after correcting for the four primary study hypotheses.

We find that our SSL pretext task of mask-and-recover performs comparably to three existing SSL pretext tasks for tabular data, namely VIME, SCARF, and SubTab, when controlling for encoder architecture with 10\% of labels across 14 tasks. The differences in performances
 are not statistically significant with Wilcoxon $p = 0.153$, $p = 0.670$, and $p = 0.502$, and non-significant after Holm-Bonferroni correction for multiple comparisons, indicating that our method cannot be statistically separated from these existing baselines when the encoders are held constant.

\section{Related Work}
\label{sec:related}

\subsection{Deep Learning for Tabular Data}

Deep learning has achieved unprecedented success on unstructured data such as images and text. Nevertheless, classical approaches based on trees, including XGBoost~\citep{xgboost}, LightGBM~\citep{lightgbm} or Random Forest~\citep{randomforest}, still prevail on tabular data. \citet{grinsztajn2022tree} suggest that deep learning performs worse in part because it is more sensitive to uninformative features and has an inductive bias less adapted to the task. Several works have emerged to develop deep learning approaches dedicated to tabular data.\pgbrk Some of them use attention to sequentially select features as in TabNet~\citep{tabnet}. Others use multi-head self-attention over tabular inputs as in FT-Transformer~\citep{fttransformer}, extend them along rows and columns as in SAINT~\citep{saint}, use contextual column embeddings as in TabTransformer~\citep{tabtransformer}, or apply attention between datapoints as in Non-Parametric Transformers~\citep{npt}. \citet{gorishniy2022embeddings} show that embeddings of numerical features play a critical role on final performance independently of the encoder used. We adopt a simpler encoder as a dual-branch MLP on categorical and numerical features. Encoder capacity is therefore the same across objectives.

\subsection{Self-Supervised Learning on Tabular Data}

Inspired by Self-Supervised Learning (SSL) successes on computer vision and natural language processing tasks~\citep{he2022mae, devlin2018bert}, recent works have studied its application to tabular data. SSL has shown promising results for deep tabular models, especially when data is scarce~\citep{vime,scarf,subtab}. VIME~\citep{vime} leverages marginal resampling as pretext task and consists of inferring corrupted features and the corruption mask. Other approaches are based on Contrastive Learning. SCARF~\citep{scarf} corrupts the data along its marginal distribution to generate positive pairs. SubTab~\citep{subtab} splits the features into subsets. It learns representations that allow predicting full rows given a subset.

\subsection{Handling Missing Data in Machine Learning}

Missing data are generally divided into three mechanisms~\citep{littlerubin}. Missingness may be completely random (MCAR), related to observed values (MAR) or also to the missing values themselves (MNAR). Classical approaches consist of imputing the data before training the final model using simple strategies like mean imputation or more sophisticated ones as K-Nearest Neighbors~\citep{troyanskaya2001} or the Multivariate Imputation by Chained Equations (MICE)~\citep{buuren2011mice}. Other methods learn how to impute the data. MissForest~\citep{stekhoven2012missforest} iteratively fills in missing features using random-forest regressors/classifiers. GAIN~\citep{yoon2018gain} uses Generative Adversarial Nets for the same purpose. Our classical baselines (Section~\ref{sec:experiments}) are centered around simple strategies to impute data, a choice we discuss in Appendix~\ref{app:baselines}.

\section{Background and Problem Setup}
\subsection{Tabular prediction with missing values}

Throughout this work, we consider supervised classification on a tabular dataset consisting of $N$ examples with $d$ features, where the $i$th example consists of numeric features $x_n^{(i)}$, categorical features $x_c^{(i)}$ and a label $y^{(i)}$ (which is known for a subset of the dataset only). To indicate missingness, we introduce a mask $m_n^{(i)} \in \{0,1\}^{d_n}$ and $m_c^{(i)} \in \{0,1\}^{d_c}$, where $d_n$ is the number of numeric and $d_c$ is the number of categorical features. If an entry is set to 1 this indicates that the respective feature is missing. To gauge dataset degradation, the numeric missingness rate is defined as follows, computed over all $N$ training examples.
\[
\text{missing\_pct} = \frac{\sum_{i=1}^{N} \sum_{j=1}^{d_n} m_{n,j}^{(i)}}{N\,d_n}\times 100\%.
\]

\subsection{Label scarcity protocol}
\label{sec:protocol}
All experiments are performed on the official train/test splits (fold 0) defined on OpenML for each dataset.
It is important to ensure that all preprocessing and pretraining is done only using the train-split to avoid seeing the test-split. Given a dataset, we sample $\{1\%,5\%,10\%,20\%\}$ labeled train sets (stratified; counts follow Table~\ref{tab:datasets}, without guaranteed class coverage). We report the mean (and, where applicable, standard deviation) across 3 seeds (0, 1, 2), controlling the sampled labeled dataset as well as fine-tuning process. When doing SSL, we only pretrain the model once and reuse this model for all 3 seeds. For the SSL-baselines and ablations in Sections~\ref{sec:ssl_baselines} and \ref{sec:ablation} respectively, we perform 3 pretraining runs, each evaluated over fine-tuning seeds 0, 1, 2.

\subsection{Metric: AUC-ROC}
We report the AUC-ROC computed on the test-split for each dataset. Evaluated across every decision threshold, binary AUC reflects the probability that a random positive case is ranked above a random negative one, with half credit for ties, where 0.5 is for random ranking and 1.0 is for perfect ranking. For the multiclass datasets in our set of 14 tasks, we compute the one-vs-rest AUC-ROC and report its macro-average over all classes. Otherwise, datasets are binary classification problems, on which we compute the standard two-class AUC-ROC.

\section{Method}
\label{sec:method}

\subsection{Model Architecture}
All the preprocessing is fit on the training data. A tabular dataset $\mathcal{D}$ is considered. For each data instance $x_i$, we denote its numerical/categorical features by $x_{n/c}$. To handle missing values, we introduce missingness indicators $m_{n/c}$, where a value of 1 indicates a missing value.
Numerical features and their corresponding missingness indicators are concatenated and processed using a dual-branch Tabular Encoder. Specifically, the numerical branch applies a two-layer MLP on the concatenated input (the categorical indicators $m_c$ do not get fed to the encoder and a missing categorical value gets treated as its own category).
$$ h_n = \text{MLP}_{\text{num}}([x_n \oplus m_n]) $$
Here $\oplus$ is the concatenation operator. We use an MLP to project $x_n$ and $m_n$ to a common hidden dimension of 128 in our experiments, applying Layer Normalization, ReLU activation function and Dropout ($p=0.1$).
Similarly, all categorical features are processed in parallel, each using a different learnable embedding dictionary $E_j \in \mathbb{R}^{|V_j| \times 32}$, where $|V_j|$ is the vocabulary size for the $j$-th categorical feature. The embeddings are then concatenated and passed through $\text{MLP}_{\text{cat}}$.
$$ h_c = \text{MLP}_{\text{cat}} \left( \bigoplus_{j=1}^{d_c} E_j(x_{c,j}) \right) $$
The final representation $h$ is obtained by combining $h_n$ and $h_c$.
$$ h = \text{MLP}_{\text{fusion}}([h_n \oplus h_c]) $$
During the pretraining phase, a decoder (a two-layer MLP for numeric features, one linear layer per categorical feature) maps $h$ back to the input features.

\subsection{Self-Supervised Pretraining Objective}
For each batch of examples, during pretraining we generate an independent artificial corruption mask per view $\tilde{m} \sim \mathrm{Bernoulli}(\rho)$, with parameter $\rho = 0.3$. For masked-out numeric features we assign $0.0$, which corresponds to the training column mean after standardization, and for masked-out categorical features we assign a special \texttt{[MASK]} token. Corruption is only applied to the available entries, and the numeric missingness indicators $m_n$ fed to the encoder are unchanged (i.e., $m_n=0$ for artificially masked entries); this choice is ablated in Section~\ref{sec:ablation}. This leads to the following ``mask, corrupt, recover'' objective. Given augmented views $\tilde{x}^{(1)}$ and $\tilde{x}^{(2)}$, we compute a consistency loss, reconstruction loss, and total SSL objective as follows.
\begin{align*}
\mathcal{L}_{\mathrm{SSL}} &= \mathcal{L}_{\mathrm{recon}}(\tilde{x}^{(1)}, x) + \mathcal{L}_{\mathrm{recon}}(\tilde{x}^{(2)}, x) + \lambda \mathcal{L}_{\mathrm{cons}}(h(\tilde{x}^{(1)}), h(\tilde{x}^{(2)}))
\end{align*}
$$ \mathcal{L}_{\mathrm{cons}} = 1 - \frac{h^{(1)} \cdot h^{(2)}}{\|h^{(1)}\|_2 \|h^{(2)}\|_2} $$
$$ \mathcal{L}_{\mathrm{num}} = \frac{\sum_{i=1}^{B}\sum_{j=1}^{d_n} \tilde{m}_{n,j}^{(i)} (\hat{x}_{n,j}^{(i)} - x_{n,j}^{(i)})^2}{\sum_{i=1}^{B}\sum_{j=1}^{d_n} \tilde{m}_{n,j}^{(i)} + \epsilon} $$
$$ \mathcal{L}_{\mathrm{cat}} = \frac{1}{|\mathcal{J}|}\sum_{j \in \mathcal{J}} \frac{\sum_{i=1}^{B} \tilde{m}_{c,j}^{(i)}\,\text{CE}(\hat{x}_{c,j}^{(i)}, x_{c,j}^{(i)})}{\sum_{i=1}^{B} \tilde{m}_{c,j}^{(i)}} $$
Here, $\mathcal{L}_{\mathrm{cons}}$ is the batch-mean cosine distance between the representations of the two views, $\mathcal{L}_{\mathrm{num}}$ is a numeric feature reconstruction loss, mean squared error computed on masked-out features, and $\mathcal{L}_{\mathrm{cat}}$ is a categorical feature reconstruction loss, mean cross-entropy computed on masked-out features. For these losses, $B$ is the batch size, $\epsilon$ is a small constant, $\mathcal{J}$ is the set of categorical features masked-out at least once and $|\mathcal{J}|$ is its size, and $\mathcal{L}_{\mathrm{recon}} = 0.5\mathcal{L}_{\mathrm{num}} + 0.5\mathcal{L}_{\mathrm{cat}}$ if both numeric and categorical features were masked-out in the current batch, and is $\mathcal{L}_{\mathrm{num}}$ or $\mathcal{L}_{\mathrm{cat}}$ if only numeric or only categorical reconstruction targets are present in a given view, respectively. If neither of them is present, the reconstruction loss is zero. We compute the mean of the cross-entropy for each feature that was masked-out and then take the mean over features; this gives each feature equal weight, as opposed to giving each masked-out token equal weight.

\subsection{Group Masking}
We compute the absolute value of the Spearman correlation matrix $|\text{corr}|$ on the numerical features and then we replace the correlations that were undefined with zero. This can be translated to a distance matrix $d = (1 - |\text{corr}|)$. We perform hierarchical clustering on this distance matrix with average-linkage. The resulting clusters are treated as groups of highly correlated features. In order to encourage learning of correlations, with probability 0.5, we additionally mask an entire cluster on top of masking individual features.

\subsection{Fine-tuning}
We pre-train the model for 50 epochs in a self-supervised manner on the unlabeled features. After pre-training, a small classification head is added and the full model is fine-tuned for 30 epochs using the labeled data with AdamW \citep{loshchilov2017decoupled} with a learning rate of $10^{-3}$, batch size 128 (or number of labeled data points if smaller), and a weight decay of $10^{-4}$.

\section{Experiments}
\label{sec:experiments}

We now study how incorporating unlabeled data during pre-training can improve downstream performance on label-scarce data with missing values.
To evaluate the effectiveness of our approach, we use 14 different classification tasks from OpenML~\citep{openml}, which are from varied domains, including four with $>1\%$ recorded numeric missingness. Detailed descriptions for the datasets can be found in Appendix~\ref{app:datasets}, and per-task missingness rates in Table~\ref{tab:main_results}.

\subsection{Baselines}
In addition to the identical model trained only on the labeled data from scratch, we compare against classical ML models trained on one 10\% labeled draw (seed 42): Logistic Regression (Log.\ Reg.\ for short), Random Forest (RF), XGBoost, and LightGBM.

We report the following three different results: Normal reports the test performance on clean test data. In order to evaluate missingness robustness we introduce missing values across available test features: MCAR randomly masks each available test feature value with probability 30\%. For MNAR shift, available numerical values exceeding the 70th percentile of their column (computed on the standardized test split, where natively missing entries are 0; ties may mask fewer than 30\%, and the categorical features are not touched) are masked.

\subsection{Statistical testing}
To evaluate statistical significance, for each task we compute the performance difference between SSL and scratch models $\Delta = \mathrm{AUC}_{\mathrm{SSL}} - \mathrm{AUC}_{\mathrm{scratch}}$. To evaluate significance across datasets, we report the mean $\Delta$ and perform a two-sided Wilcoxon signed-rank test as suggested by \citet{demsar2006}. In our main experiments we have four hypothesis tests, label scarcity (5\%, 10\%), robustness to MCAR and robustness to MNAR shift. We correct for this family by Holm-Bonferroni correction at $\alpha = 0.05$~\citep{holm1979}. The exploratory tests done at 1\% and 20\% labels gave us $p=0.463$ and $0.542$. Throughout the paper, we report Holm-adjusted as well as original $p$-values for the primary tests and the SSL-baseline comparison but the ablation $p$-values in Section~\ref{sec:ablation} are uncorrected. In Section~\ref{sec:ssl_baselines}, we compare our method to previous tabular SSL approaches; this constitutes another family of 3 hypothesis tests that are corrected accordingly.

\section{Results}\label{sec:results}

\subsection{Main results under label scarcity}
\label{sec:main_results}
\begin{figure}[H]
  \centering
  \includegraphics[width=0.7\linewidth]{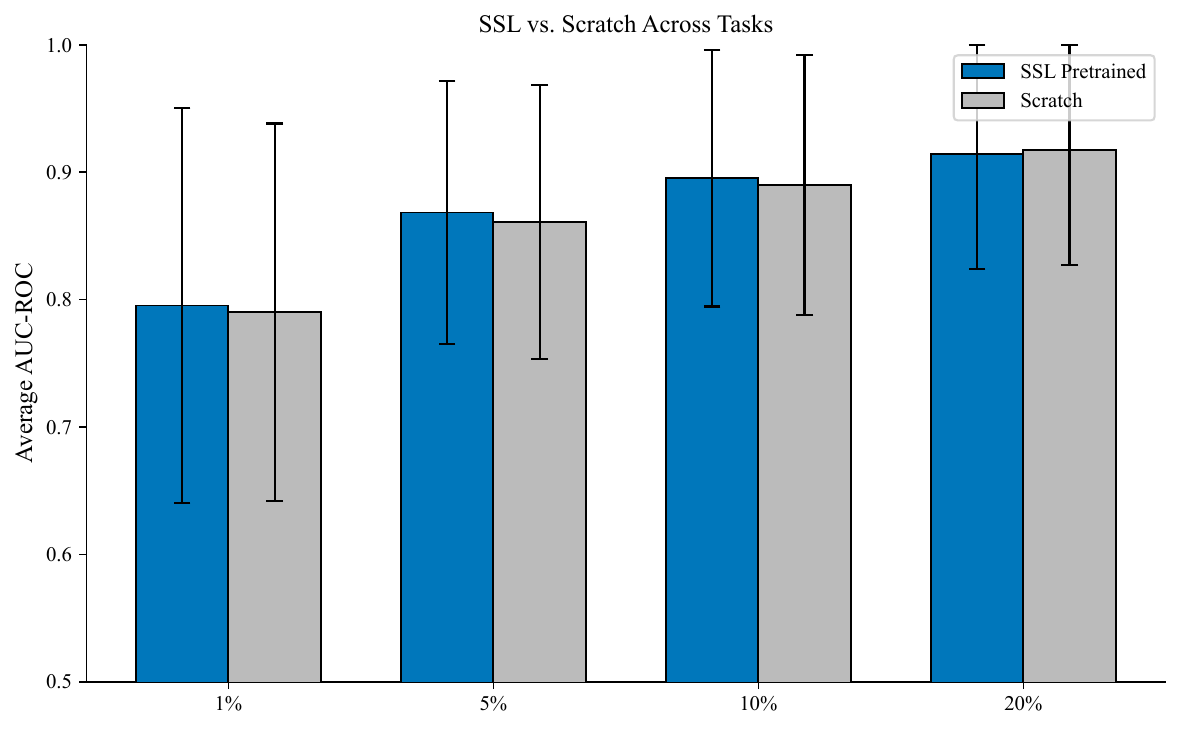}
  \setlength{\abovecaptionskip}{2pt}\caption{Mean test AUC-ROC across tasks (SD across tasks; y-axis from 0.5; clipped at 1.0).}
  \label{fig:main_bar}
  \vspace{-4pt}
\end{figure}

\looseness=-1 In Figure~\ref{fig:main_bar}, we show the averaged AUC-ROC scores across the different datasets.
With only 5\% labels, SSL provides an improvement over training from scratch with an average AUC improvement of $+0.0074$. With 10\% labeled data the average AUC improvement is $+0.0055$.

\noindent
However, aggregating the performance improvements with a Wilcoxon signed-rank test does not detect a significant difference for either 5\% or 10\% labels ($p = 0.6257$; Holm-adjusted $p = 1.000$). The per-task results show that pretraining with SSL is not necessarily helpful on all datasets and SSL methods should not be expected to help on every task.\footnote{Regarding the equal $p$-values for different fractions: different paired differences can give the same exact Wilcoxon $p$-value; having equal $p$-values does not suggest a bug by itself.}
Also, for 20\% labeled data the trend is even reversed with SSL performing worse than models trained from scratch (avg.\ AUC SSL $0.9145$ compared to avg.\ AUC scratch $0.9173$), weakly suggesting (none of the fractions are significant) that potential benefits from SSL are concentrated in \mbox{settings} with little labeled data, but do not necessarily continue when more labeled data becomes available. Per-task results are in \mbox{Appendix~\ref{app:per_task}}.

\subsection{The Missing vs. Clean Paradox}
\label{sec:missing_clean}
\begin{figure}[H]
  \centering
  \includegraphics[width=0.75\linewidth]{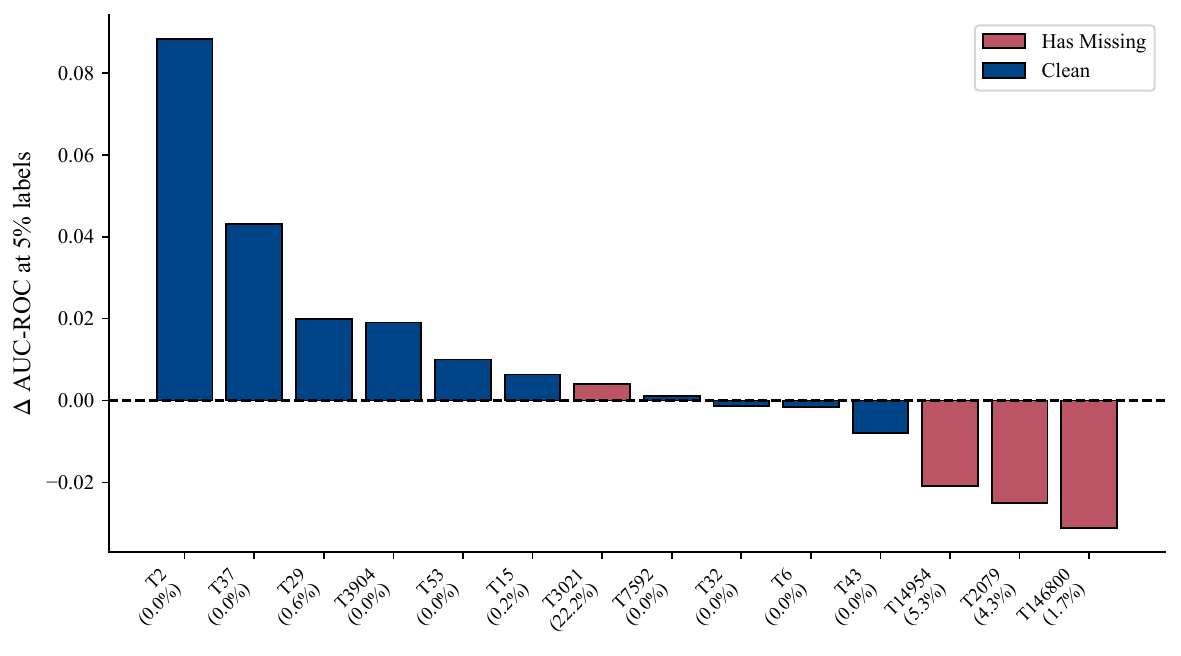}
  \setlength{\abovecaptionskip}{2pt}\caption{Per-task $\Delta$AUC (SSL$-$Scratch) at 5\% labels. Datasets with $>1\%$ native numeric missingness are ``Has Missing''; the rest are ``Clean.'' (The 1\% threshold was chosen post hoc.)}
  \label{fig:missing_vs_clean}
\end{figure}

A major impetus behind mask-and-recover SSL methods is the idea that SSL could potentially help models learn representations which deal with missingness natively. In practice, we found SSL's improvements to be most consistent for the numerically clean data. On datasets with inherent missingness, SSL had highly variable effects and often reduced performance (SSL wins on 7 of 10 clean tasks at 5\% labels but only 1 of 4 tasks with missing data; descriptive, untested; one pretraining run per task). These groups differ in other ways too, so this is not causal. See Figure~\ref{fig:missing_vs_clean}.

\subsection{Robustness to Test-Time Degradation}
\label{sec:robustness}
\begin{figure}[htbp]
  \vspace{8pt}
  \centering
  \includegraphics[width=0.6\linewidth]{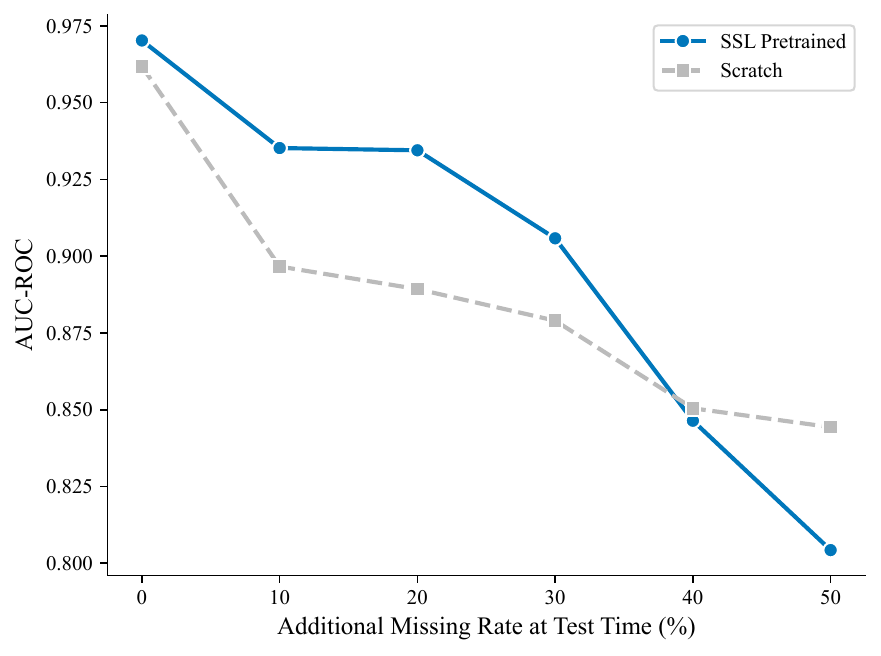}
  \caption{AUC-ROC on Task 3021 as additional MCAR missingness is injected at test time (single seed).}
  \label{fig:mcar_curve_3021}
\end{figure}

SSL-pretrained models have higher average raw AUC on corrupted test data compared to their counterparts trained from scratch (10\% labels, single fine-tuning seed, one fixed MCAR draw). Specifically, SSL improves AUC by $+0.0245$ on the MCAR $+30\%$ numeric/categorical features test compared to scratch (raw $p = 0.0295$; Holm-adjusted $p = 0.118$ across our family of four primary tests). Similarly, SSL improves performance in the presence of the MNAR missingness shift, increasing performance by $+0.0418$ AUC on average compared to scratch (raw $p = 0.1726$; Holm-adjusted $p = 0.518$). While not statistically significant after family-wise correction, these are actually our two largest effect sizes of the four primary tests, but note that Task 2 (Anneal) influences both in a substantial way (without it: $+0.0144$ and $+0.0300$). See Figure~\ref{fig:mcar_curve_3021} for how AUC changes as missingness increases on Task 3021 (single seed; the task with the most natively missing values, mostly from one empty column).

\FloatBarrier
\vspace{-15pt}

\subsection{Ablation study}
\label{sec:ablation}
\flushbottom
\newlength{\savedparskip}\setlength{\savedparskip}{\parskip}
\setlength{\parskip}{0pt plus 8pt}

\begin{figure}[!htbp]
  \centering
  \includegraphics[width=0.95\linewidth]{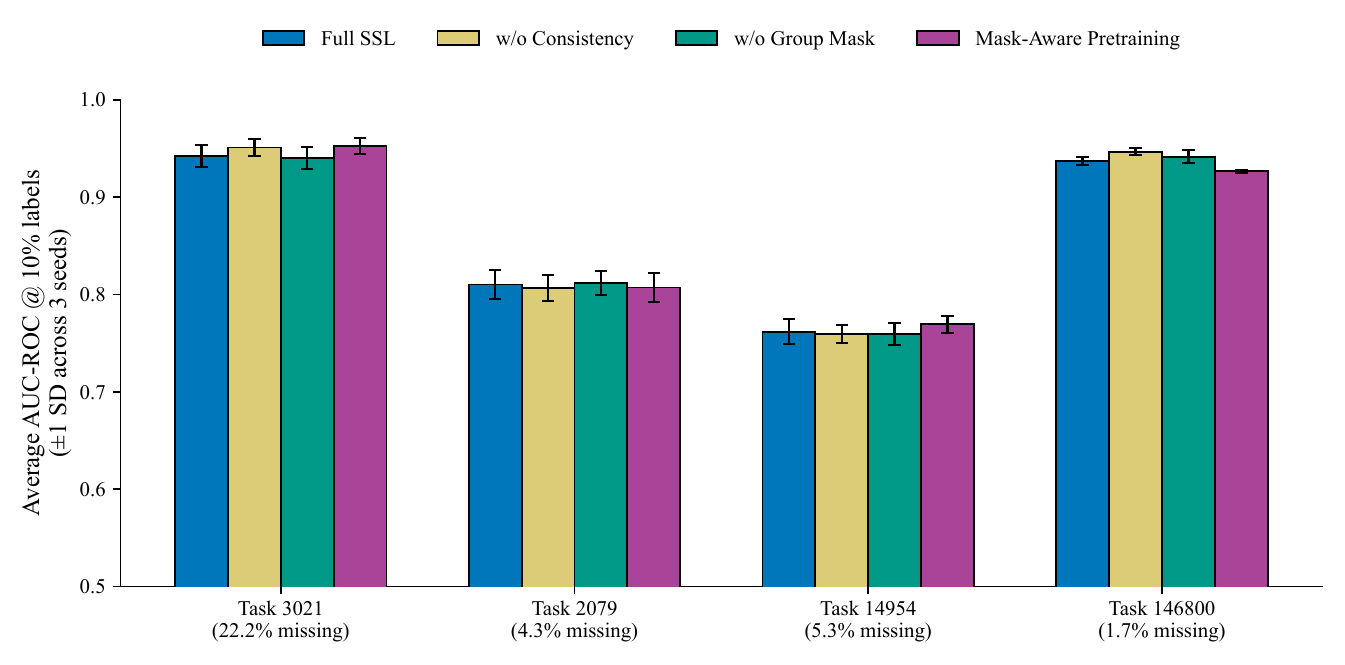}
  \caption{Ablation results at 10\% labels across the four native-missingness tasks (missingness \% in parentheses). Results are averaged over 3 pretraining seeds, each over 3 fine-tuning seeds; error bars are $\pm 1$ sd over pretraining seeds (y-axis starts at 0.5; upper bars clipped at 1.0). ``Mask-Aware'' denotes using the pretext corruption mask for pretraining as input to the encoder via $m$.}
  \label{fig:ablation}
\end{figure}

Next, we ablate the components that could have particular relevance in the presence of missingness (consistency loss, group masking and mask-visibility) and evaluate performance at 10\% labels (Figure~\ref{fig:ablation}). Here, we look at tasks with non-trivial missingness natively (Task 3021, 22.2\% missing (mostly one empty column), Task 14954, 5.3\% missing, Task 2079, 4.3\% missing, Task 146800, 1.7\% missing). From these four tasks, a single pretraining seed could suggest that ablating consistency loss improved performance, but the per-pretraining-seed mean effect, which was averaged over the four tasks, ranged from $-0.0004$ to $+0.0064$.

However, when repeating this experiment
with $n=3$ seeds and averaging the AUC-ROC scores (of the 3 seeds) before taking the difference (variant minus full model; positive values favor the variant), ablating consistency loss does not reliably improve performance: mean $\Delta$AUC: $+0.0031$(p-value: $0.625$, Wilcoxon signed-rank test for the 4 task-level mean values with $n=4$; p-value: $0.233$, Wilcoxon signed-rank test for the $3 \times 4$ pretraining-seed means with $n=12$ ; Task-wise $\Delta$AUCs: $+0.0088$ for Task 3021, $-0.0025$ for Task 14954, $-0.0037$ for Task 2079, $+0.0098$ for Task 146800), suggesting that the effects of ablating the consistency loss seen for the individual (single-seed) pretraining model vary across the seeds/initializations of the pretraining phase. Ablating group masking has a mean $\Delta$AUC of $+0.0004$ (p-value: $1.0$; Task-wise $\Delta$AUCs: $-0.0022$ for Task 3021, $-0.0024$ for Task 14954, $+0.0016$ for Task 2079, $+0.0045$ for Task 146800). In general, we find per-task standard deviations for the full model on these four tasks between 0.004 and 0.015 AUC (population SD, $n=3$) across pretraining seeds, indicating that effects that seem consistent across tasks for a single-seed may partly be due to randomness in the pretraining seed/initialization.

In the corruption pipeline described in Section~\ref{sec:method}, artificial masking is done but the mask indicator is left at $m{=}0$. We rerun the pretraining procedure recording the pretext masking in $m$. We term this approach ``Mask-Aware''; since the encoder ignores $m_c$, only numeric mask visibility changes. We conduct the experiment with 3 random initialisation seeds. We examined whether the Missing vs.\ Clean Paradox (Section~\ref{sec:missing_clean}) happens because the encoder never sees entries that are artificially masked as $(x{=}0, m{=}1)$, which is how actual missingness manifests. However, this effect proved inconsistent across the four tasks (mean $\Delta$AUC $= +0.0011$, $p=0.875$: Task 3021 $+0.0105$, Task 2079 $-0.0031$, Task 14954 $+0.0074$, Task 146800 $-0.0104$), and this does not test the between-group gap, since only higher-missingness tasks were included. All three components for which we conducted ablations had no effect that could be detected (low power) after averaging over random seeds for pretraining.

This is a valuable insight into the difficulty of conducting ablations in the setting of tabular SSL. Specifically, averaging over random seeds is necessary before we make conclusions on the effects of our design decisions on performance. Patterns can be misleadingly apparent in setups where performance is assessed with a single seed since variance can be high and there are a comparatively low number of tasks, in our case 4 native-missingness tasks.
\enlargethispage{1.5\baselineskip}
\vspace{0pt plus 80pt}
\pagebreak[4]
\setlength{\parskip}{\savedparskip}

\subsection{Classical baselines at 10\% labels}
\label{sec:baselines}

As seen in Table~\ref{tab:baselines}, our deep model, with or without any SSL pretraining has a performance competitive with the classical baselines in the case where only 10\% of labels are available. Our SSL model performs best in comparison to the other baselines except Random Forest, with a difference of $+0.0688$ AUC from Logistic Regression, $+0.0135$ AUC from XGBoost and $+0.0079$ AUC from LightGBM. For the classical baselines mean imputation, ordinal encoding of categoricals, and no feature scaling are used (so LightGBM and XGBoost do not utilize the native NaN handling that they have). However, the neural models get the standardized inputs, so the difference for Logistic Regression may be partly a preprocessing artifact (unscaled and below chance on Task 14954) (Appendix~\ref{app:baselines}). Classical models also mode-impute categoricals, unlike the neural models (a confound on Anneal).
\vspace{0pt}
\begin{table}[htbp]
\centering
\small
\begin{tabular}{lc}
\toprule
Method & Avg AUC @ 10\% labels \\
\midrule
Random Forest         & \textbf{0.9015} \\
SSL Pretrained (Ours) & 0.8954 \\
Scratch (No Pretrain) & 0.8899 \\
LightGBM              & 0.8875 \\
XGBoost               & 0.8819 \\
Logistic Regression   & 0.8266 \\
\bottomrule
\end{tabular}
\caption{Baseline comparison averaged across tasks at 10\% labels. Classical methods are trained with one fixed seed (42; a different labeled subset than the neural models). SSL Pretrained and Scratch methods are trained with one shared pretraining (SSL only) and 3 independent fine-tuning seeds.}
\label{tab:baselines}
\end{table}
\vspace{-10pt}
\enlargethispage{1.5\baselineskip}

\begin{table}[!b]
\vspace{0pt}
\centering
\fontsize{10.5}{12.5}\selectfont
\renewcommand{\arraystretch}{1.1}
\begin{tabular}{lccc}
\toprule
Method & Avg AUC @ 10\% & Ours $-$ baseline & Adj. $p$ \\
\midrule
Ours (mask-and-recover) & 0.8954 & --- & --- \\
VIME \citep{vime} & 0.8978 & $-0.0024$ & 0.459 \\
SCARF \citep{scarf} & 0.8943 & $+0.0011$ & 1.000 \\
SubTab \citep{subtab} & 0.8913 & $+0.0041$ & 1.000 \\
\bottomrule
\end{tabular}
\caption{Comparison against simplified prior tabular SSL implementations using an identical encoder architecture. $\Delta$ is defined as Ours $-$ baseline; adjusted $p$-values are Holm-Bonferroni corrected for a family of 3. Ours does one run for pretraining; baselines get an average of 3.}
\vspace{+12pt}
\label{tab:ssl_baselines}
\end{table}

\vspace{-10pt}
\subsection{Comparison against prior tabular SSL objectives}
\label{sec:ssl_baselines}
To evaluate our mask-and-recover objective in an architecture-matched comparison with existing tabular SSL approaches, we implement simplified versions of VIME~\citep{vime}, SCARF~\citep{scarf} and SubTab~\citep{subtab} using the same model architecture (that is, the two-branch encoder described in Section~\ref{sec:method}), though optimization and corruption settings differ. Results are given in Table~\ref{tab:ssl_baselines}, where we conduct experiments on all 14 tasks with only 10\% labels, 
and average the baselines' results over 3 pretraining seeds; the numbers for our method are the main-results SSL numbers (Table~\ref{tab:main_results}; one shared pretraining run, 3 fine-tuning seeds). 
\pagebreak

\noindent We implemented all three baselines in our own code base (VIME without its semi-supervised stage). 
Other than the pretraining objective, they differ from our setup as follows: batch size 128 (ours: 256), constant learning rate (ours: OneCycleLR), and no gradient clipping (ours: 1.0). 
The optimizer (AdamW, learning rate $10^{-3}$, weight decay $10^{-4}$), the 50 pretraining epochs and the fine-tuning procedure are the same but the smaller batch size gives them about 2$\times$ more steps on every task (2--4$\times$ for VIME and 2--3$\times$ for SCARF and SubTab on smaller tasks). 
The remaining settings for this are in Appendix~\ref{app:hyperparameters}.
\noindent None of the methods show a statistically significant difference to ours under a Wilcoxon test, namely VIME (mean
 $\Delta = -0.0024$, $p = 0.153$), SCARF (mean $\Delta = +0.0011$, $p = 0.670$), and SubTab (mean $\Delta = +0.0041$, $p = 0.502$), with all remaining non-significant after Holm-Bonferroni adjustment (adjusted $p = 0.459$, $1.000$, and $1.000$, respectively), though VIME's mean is mildly above ours. We do not draw conclusions from this about the irrelevance of different SSL objectives given the limited number of experiments; rather, these results indicate that our model performs competitively (not equivalence-tested) with the reimplemented SSL objectives for tabular data. Further analysis of SSL objectives for tabular data with larger-scale experiments may be interesting for future work.

\section{Discussion}
\paragraph{When does SSL help the most?}
SSL seems to have the largest gains at 5\% labels, though none of the four fractions is significant (the 5\% gain is $+0.0012$ without Task 2), and the average gain turns negative at 20\%. SSL also helps the most consistently on datasets that are numerically clean (though the biggest winner, which was Task 2, had 77.2\% missing/not-applicable categorical values, and Task 37, which was second, appears to code missing values as zeros and without both, the mean of the clean group $\Delta$ at 5\% goes from $+0.0177$ to $+0.0057$), which goes against the intuition that SSL's ability to handle missing values would be helpful on datasets with lots of missingness. The clean minus missing gap has the same sign at all four of the label fractions.

\paragraph{Why might SSL have higher AUC under test-time missingness?}
These are corrupted-test AUCs, not degradation measures, and the explanation below is untested without a scratch control with masking augmentation. Under MCAR, scratch gets a categorical \texttt{[MASK]} embedding too, without any data dependent training. We hypothesize that this may be because mask-and-recover pretraining encourages the encoder to make use of redundant information present among features, which can then make the model more robust to features being missing at test time. We do find that SSL shows a higher average corrupted-test AUC, with a  $+0.0418$ AUC mean shift under structured MNAR missingness shift and positive impact on 8/14 tasks (median $+0.0066$; mean $+0.0018$ not counting the four largest-delta tasks 2, 3904, 37 and 43), but not on three of the four native-missingness tasks. Note that on Task 3021 the SSL  benefit is not present beyond 30\% missingness (Figure~\ref{fig:mcar_curve_3021}), where SSL's performance scores below scratch at 40\% and 50\%, in this one-task, one-seed evaluation. However, this should be taken with a grain of salt as this curve was not tested; the MNAR comparison above has adjusted $p=0.518$.

\paragraph{Why are some components not always helpful?}
Lastly, SSL has many knobs to tweak, and while ablating each of them can be insightful, it is critical to average over many seeds before drawing conclusions, especially in a low-sample regime like ours. For instance, from just one seed, it would be easy to conclude that consistency regularization hurts on datasets with natural missingness, but when averaged over three pretraining seeds for each SSL variant in the ablation in Section~\ref{sec:ablation}, the impact of consistency regularization, group masking, and mask-aware pretraining has different signs across tasks and is not distinguishable from zero on aggregate. More tasks or pretraining seeds may be required to draw conclusions about these minor effects.

\section{Limitations and Broader Impact}
\label{sec:limitations}
\begingroup

\setlength{\parskip}{0pt plus 8pt}

\paragraph{Limitations.} 
Although our SSL method achieves a higher average AUC in conditions of test-time missingness, there are a number of limitations. First, our mask-and-recover pretraining task does not always help the learning algorithm, for reasons that are not established here. For instance, introducing artificial missingness on tabular data which already have significant missingness may lead the model to underfit the data.
\noindent Second, running 50 pretraining epochs and 30 fine-tuning epochs may demand greater computational expense than fitting classical ML approaches like decision trees (runtime was not measured). Third, the feature-grouping strategy relies on Spearman rank correlation between features, which may fail to capture other kinds of relations. Finally, although we report the mean across 3 random seeds for most of the deep-learning results (label-scarcity experiments with one pretraining run which was shared, ablations, comparison with SSL baselines), the results in Section~\ref{sec:robustness} about MCAR and MNAR missingness (the ``$\Delta$ MCAR'' and ``$\Delta$ Shift'' columns of Table~\ref{tab:robustness}) were computed by fixing the final fine-tuning random seed (42), so these numbers may be noisy with respect to seed choice. Moreover, under MCAR, masked categorical entries use the \texttt{[MASK]} token, which has only been seen by the pretrained encoder, and the MCAR and MNAR rates (up to $30\%$) are equal to the nominal pretraining masking rate (before group masking), possibly favoring SSL; also there isn't any scratch control with masking augmentation. The injected entries have $m{=}1$ but the pretraining masks have $m{=}0$; for features without natively missing entries, these $m$-weights receive no training signal, so the injections test an untrained input channel too. For seed 42, the mean of the ``$\Delta$ 10\%'' column ($+0.0086$) is different from the $+0.0055$ reported in Section~\ref{sec:main_results}, which was computed by averaging 3 different random seeds. Also, the single shared pretraining run can change the per-task AUC score by roughly 0.004--0.015 (the pretraining-seed SD on the four ablation tasks), which is comparable to or larger than the mean SSL gain. Relatedly, for comparison with classical methods in Table~\ref{tab:baselines}, we report the result of a fixed random seed (see Appendix~\ref{app:baselines}) and did not show the variance across random seeds. Lastly, although our paired statistical tests nominally include $n = 14$ tasks, performance tends to be saturated for some of the tasks. In particular, Tasks 15 and 32 already achieve (or exceed) an AUC of about $0.98$ from 1\% labels onward. Task 6 already achieves (or exceeds) an AUC of about $0.98$ from 5\% labels onward. For Task 6, however, both SSL and scratch achieve an AUC below $0.93$ with 1\% labels. Differences among methods are therefore relatively small and low-variance for these three tasks, as ceiling effects leave little room for gains. Seven test sets are also small (54--90 examples; Table~\ref{tab:datasets}). Eight tasks receive only 50--150 pretraining updates, so training budget differs with task characteristics; three of the four tasks which have higher missingness are in this group. We did not verify subject-level separation for repeated measurements such as MiceProtein.

\noindent Our comparisons of robustness are not significant after correction. Therefore, improvements, e.g., $+0.0245$ AUC for MCAR, should be taken as suggestive evidence rather than proof. Our ablation study suffers from limited statistical power. It was conducted on only four tasks (those with native-missingness) and we computed the average of the results of three independent pretraining seeds for each variant. We could not achieve $p < 0.05$ on the resulting Wilcoxon test on four pairs, no matter how strong the effect (the $n=12$ test treats the task--seed pairs as independent pairs, so it is only exploratory). Thus, we encourage the reader to take our null findings for consistency loss, group masking, and mask-aware components as no evidence of an effect, not evidence of no effect. 
\par\noindent We only compared to other SSL methods at 10\% labels, in Section~\ref{sec:ssl_baselines}. Other percentages may result in different findings. We did not perform any hyperparameter optimization for any given dataset and method combination (see Appendix~\ref{app:hyperparameters}). Doing so may result in different findings, including for the Missing vs.\ Clean Paradox. This choice avoided dataset-specific tuning, though it does not isolate the objective and limits how broadly these comparisons generalize. While we explored whether mask-visibility in pretraining could explain the Missing vs.\ Clean Paradox in Section~\ref{sec:ablation}, we were not able to say whether it does. We encourage the reader to consider the root-cause of the paradox an open question. 
Finally, we report raw and Holm-adjusted $p$-values for both the primary tests and the SSL-baseline comparison with their correction applied within their families (Section~\ref{sec:experiments}); the $p$-values of the ablation analyses are raw and uncorrected. 
Raw effect sizes (e.g., the $+0.0245$ AUC gain under MCAR) should be read as suggestive rather than confirmatory, because none of the primary robustness comparisons remain significant after correction.

\paragraph{Broader Impact.} A potential benefit is robustness to label scarcity: it is of great value to train a robust model when labels require costly domain expertise, e.g., in medical diagnostics or fraud detection. In addition, if achieved, robustness to sensor degradation (i.e., MNAR test-time shift, an effect which was driven mainly by a small number of tasks) could reduce the risk of catastrophic model failure upon deployment. That being said, practitioners should be cautious when using deep tabular models in high-stakes applications. Deep tabular models are less interpretable by nature compared to shallow decision trees, and therefore, should be used with additional explainability checks to ensure that the model is not learning spurious correlations.

\endgroup
\section{Conclusion}
Overall, we found that mask-and-recover SSL can show a small average advantage over training from scratch, and performs slightly below Random Forest and above the default XGBoost and LightGBM (one draw for the classical models; but comparisons untested). However, SSL is a complex and highly dataset-dependent endeavour in the context of tabular data, with substantial inter-task variance in the extent to which it can offer gains. Moreover, our controlled comparisons do not always paint a uniformly rosy picture for SSL, including: (i) a paradoxical ``clean vs. missing'' pattern: the largest improvements were found on datasets with low amounts of numeric missingness (post hoc, descriptive); (ii) seed-averaged ablations indicating no detectable impact of view-consistency regularization, group masking, or mask-aware pretraining on tasks with native missingness. Notably, we do not reliably replicate an initial (single-seed) impression that view-consistency losses hurt performance. We recommend practitioners to carefully evaluate the effect of pretraining across multiple random seeds before drawing conclusions, particularly in the context of small-sample studies in SSL for tabular data. In particular, we find that, averaged over diverse OpenML datasets, SSL may offer small, task-dependent gains over scratch in scenarios of (a) scarce labelled data (though the gains at 5\% and 10\% labels are not statistically significant, $p = 0.6257$) and (b) test-time missingness: SSL-pretrained models both achieve a higher average AUC on corrupted test data, averaged over all 14 tasks, under random (MCAR) missingness injections and structured (MNAR) test-time shifts compared to their counterparts trained from scratch, though neither difference remains statistically significant (after correction for multiple comparisons). 
\par\noindent Furthermore, our mask-and-recover strategy showed no significant difference from our reimplementations of VIME~\citep{vime}, SCARF~\citep{scarf}, and SubTab~\citep{subtab}, under an identical architecture (not equivalence-tested; a single pretraining run is used for ours versus a 3-run average for the baselines).

\section{Future Work}
Although we were not able to explain the Missing vs.\ Clean Paradox with our mask-aware ablation, its underlying cause remains an open question for future research. For instance, since native and synthetic missingness are not identical, one might explore corruption schemes that take the native missingness distribution into account in some way, e.g., by modeling native and synthetic missingness jointly.
Our ablations with seed averaging indicated that neither consistency loss nor group masking had a consistent effect on the four native-missingness tasks that we considered. It would be interesting to conduct a similar ablation across an expanded and more varied collection of native-missingness datasets, in order to have the power to tell if these techniques offer a small, genuine effect for missingness, or if they truly do not help with this task.
We limited our study which focuses on missingness to mask-and-recover pretraining, so it would be interesting to repeat this analysis for other types of SSL pretraining (e.g., contrastive or generative) to determine whether any of the benefits or paradoxical behaviors that we observed in this work are unique to mask-and-recover pretraining or generalize to SSL for tables in general. Finally, we hope that the use of seed averaging in this paper will prove useful in future empirical studies of representation learning for tabular data.
\par
\section*{Declarations}
\textbf{Funding.} This work has been conducted without any external funding. \\[4pt]
\textbf{Conflicts of Interest.} The author has declared that there are no competing interests. \\[4pt]
\textbf{Data Availability.} The data used in this paper are all available from the OpenML repository. \\[4pt]
\textbf{Code Availability.} All of the PyTorch source code, training, and evaluation scripts used to produce the findings reported in this paper are located at \url{https://github.com/shnd23/ssl-tabular-label-scarcity}. \\[4pt]
\textbf{Acknowledgments.} The breast cancer database (Task 15) is from Dr. William H. Wolberg, University of Wisconsin Hospitals, Madison.

\clearpage
\appendix

\section{Per-task Results and Robustness Table}
\label{app:per_task}

Tables in this appendix show the exact evaluation scores for each task for the aggregated results in the main text. Table~\ref{tab:main_results} shows the test AUC-ROC scores for the Self-Supervised Learning (SSL) pretrained encoder and Scratch baseline. Results are shown for each of the label scarcity settings (1\%, 5\%, 10\%, 20\%). Table~\ref{tab:robustness} shows the exact $\Delta$ AUC (SSL$-$Scratch) for the normal (i.e.\ 10\% labelled) setting. Results are shown for each of the test-time degraded settings, MCAR $+30\%$ (Missing Completely At Random) and structured distributional shift (MNAR: Missing Not At Random).

\begin{table}[!htbp]
\centering

\resizebox{1\textwidth}{!}{%
\begin{tabular}{rrrrrrrrrr}
\toprule
\textbf{Task} & \textbf{Num. missing(\%)} & \textbf{SSL 1\%} & \textbf{Scr 1\%} & \textbf{SSL 5\%} & \textbf{Scr 5\%} & \textbf{SSL 10\%} & \textbf{Scr 10\%} & \textbf{SSL 20\%} & \textbf{Scr 20\%} \\
\midrule
2079 & 4.305 & 0.554 & 0.560 & 0.742 & 0.767 & 0.796 & 0.807 & 0.848 & 0.869 \\
3021 & 22.216 & 0.938 & 0.872 & 0.934 & 0.930 & 0.957 & 0.948 & 0.970 & 0.970 \\
14954 & 5.258 & 0.594 & 0.680 & 0.738 & 0.759 & 0.780 & 0.806 & 0.799 & 0.830 \\
146800 & 1.732 & 0.688 & 0.685 & 0.881 & 0.912 & 0.943 & 0.956 & 0.987 & 0.992 \\
2 & 0.000 & 0.700 & 0.594 & 0.864 & 0.776 & 0.976 & 0.923 & 0.991 & 0.982 \\
15 & 0.247 & 0.993 & 0.988 & 0.991 & 0.985 & 0.987 & 0.989 & 0.987 & 0.983 \\
29 & 0.617 & 0.843 & 0.858 & 0.866 & 0.846 & 0.866 & 0.842 & 0.885 & 0.887 \\
3904 & 0.010 & 0.662 & 0.640 & 0.643 & 0.624 & 0.650 & 0.633 & 0.690 & 0.672 \\
7592 & 0.000 & 0.873 & 0.858 & 0.882 & 0.881 & 0.885 & 0.886 & 0.893 & 0.895 \\
6 & 0.000 & 0.925 & 0.923 & 0.981 & 0.982 & 0.994 & 0.994 & 0.998 & 0.998 \\
32 & 0.000 & 0.989 & 0.987 & 0.997 & 0.999 & 1.000 & 1.000 & 1.000 & 1.000 \\
37 & 0.000 & 0.835 & 0.832 & 0.859 & 0.816 & 0.840 & 0.805 & 0.881 & 0.875 \\
43 & 0.000 & 0.929 & 0.924 & 0.938 & 0.946 & 0.943 & 0.953 & 0.953 & 0.965 \\
53 & 0.000 & 0.614 & 0.661 & 0.841 & 0.831 & 0.919 & 0.916 & 0.921 & 0.924 \\
\bottomrule
\end{tabular}%
}
\caption{Per-task AUC-ROC results for SSL pretraining vs. scratch training across four label scarcity fractions. Metrics represent the mean over 3 random seeds and are rounded to 3 decimals for readability; statistical tests use full-precision values, available in the accompanying code. SSL values use one pretraining run which was shared, so they are different from the three seed pretraining ``Full'' model in the ablation (Figure~\ref{fig:ablation}).}
\label{tab:main_results}
\end{table}

\enlargethispage{2\baselineskip}
\begin{table}[H]
\vspace{-35pt}
\centering
\fontsize{9}{10}\selectfont
\setlength{\tabcolsep}{10pt}
\renewcommand{\arraystretch}{0.9}
\begin{tabular}{rrrrr}
\toprule
\textbf{Task} & \textbf{Num. missing(\%)} & \textbf{$\Delta$ 10\%} & \textbf{$\Delta$ MCAR} & \textbf{$\Delta$ Shift} \\
\midrule
2079 & 4.3051 & -0.0071 & -0.0223 & -0.0113 \\
3021 & 22.2157 & 0.0085 & 0.0267 & -0.0185 \\
14954 & 5.2583 & -0.0456 & -0.0000 & 0.0063 \\
146800 & 1.7316 & -0.0133 & -0.0203 & -0.0466 \\
2 & 0.0000 & 0.1100 & 0.1551 & 0.1954 \\
15 & 0.2473 & 0.0109 & 0.0154 & -0.0018 \\
29 & 0.6173 & -0.0026 & 0.0051 & 0.0068 \\
3904 & 0.0097 & 0.0465 & 0.0418 & 0.1661 \\
7592 & 0.0000 & -0.0024 & 0.0117 & -0.0014 \\
6 & 0.0000 & 0.0007 & 0.0006 & -0.0210 \\
32 & 0.0000 & -0.0001 & 0.0107 & 0.0347 \\
37 & 0.0000 & 0.0193 & 0.0548 & 0.1052 \\
43 & 0.0000 & -0.0081 & 0.0160 & 0.0997 \\
53 & 0.0000 & 0.0039 & 0.0475 & 0.0713 \\
\bottomrule
\end{tabular}
\caption{Robustness deltas at 10\% labels (seed 42). Values represent $\Delta$AUC (SSL$-$Scratch) under normal, MCAR (+30\% missingness), and MNAR structured shift. For Task 14954, the MCAR delta is $-1.1\times10^{-16}$, a numerical tie that was treated as a nonzero negative difference by the Wilcoxon test, and dropping it gives raw $p=0.0327$ (Holm-adjusted $\approx0.131$).}
\label{tab:robustness}
\end{table}
\FloatBarrier

\section{Data Dictionary and Context}
\label{app:datasets}
To test our results across distinct data, we have curated a set of 14 data sets from the OpenML repository (this selection was not tuned on the results) that vary in domains, number of features, and other characteristics including the amount of missingness. Table~\ref{tab:datasets} shows each task's size. No dataset was dropped after seeing results (only the tasks with fewer than 300 training rows are skipped by the code). We include descriptions of each below and we cite the original source of each of the datasets where it is requested; all of them were retrieved from OpenML~\citep{openml}:
\vspace{10pt}
\begin{table}[htbp]
\centering
\small
\setlength{\tabcolsep}{7pt}
\begin{tabular}{rlrrrrr}
\toprule
\textbf{Task} & \textbf{Dataset} & \textbf{$n_{\text{train}}$} & \textbf{$n_{\text{test}}$} & \textbf{Features} & \textbf{Classes} & \textbf{Labels @ 1\%} \\
\midrule
2079 & Eucalyptus & 662 & 74 & 19 & 5 & 6 \\
3021 & Sick & 3394 & 378 & 29 & 2 & 33 \\
14954 & Cylinder Bands & 486 & 54 & 37 & 2 & 4 \\
146800 & MiceProtein & 972 & 108 & 77 & 8 & 9 \\
2 & Anneal & 808 & 90 & 38 & 5 & 8 \\
15 & Breast-W & 629 & 70 & 9 & 2 & 6 \\
29 & Credit-Approval & 621 & 69 & 15 & 2 & 6 \\
3904 & JM1 & 9796 & 1089 & 21 & 2 & 97 \\
7592 & Adult & 43957 & 4885 & 14 & 2 & 439 \\
6 & Letter & 18000 & 2000 & 16 & 26 & 180 \\
32 & Pendigits & 9892 & 1100 & 16 & 10 & 98 \\
37 & Diabetes & 691 & 77 & 8 & 2 & 6 \\
43 & Spambase & 4140 & 461 & 57 & 2 & 41 \\
53 & Vehicle & 761 & 85 & 18 & 4 & 7 \\
\bottomrule
\end{tabular}
\caption{The dataset sizes for fold 0 of the 10-fold split of each task on OpenML. Labels at 1\% is $\max(\lfloor 0.01\,n_{\text{train}}\rfloor,\ \text{number of classes})$, so eight of the tasks have fewer than 10 labeled examples at 1\%.}
\label{tab:datasets}
\end{table}

\textbf{Task 2079 (Eucalyptus):} An ecological data set to predict the utility of Eucalyptus seedlots used for soil conservation. Has both categorical features (describing site and species attributes) and numerical features, and contains 4.3\% numeric missingness .

\textbf{Task 3021 (Sick):} A medical data set for thyroid disease diagnosis that contains 22.2\% numeric missingness, mainly because the TBG column is completely empty (the other numeric columns have the average of 9.2\%)~\citep{quinlan1987sick}.

\textbf{Task 14954 (Cylinder Bands):} An industrial data set for predicting if banding which is a cause of delays, happens during the rotogravure printing press process. Has both nominal and numerical features, with 5.3\% numeric missingness~\citep{evans1994cylinder}.

\textbf{Task 146800 (MiceProtein):} A biological data set that measures the abundance of 77 proteins in the cerebral cortex of mice. Contains 1.7\% numeric missingness and can be used to identify subpopulations of mice exposed to particular learning contexts~\citep{higuera2015}.

\textbf{Task 2 (Anneal):} A metallurgical data set for predicting the steel annealing process that has many missing values in the original (categorical) version of the dataset. We use the OpenML version of the data set, which contains 0.0\% missing numeric values but 77.2\% missing/not-applicable categorical values (averaged over the categorical columns)~\citep{annealing_uci}.

\textbf{Task 15 (Breast-W):} Also known as Wisconsin Breast Cancer data set. Uses nine cytological characteristics that are rated from 1--10~\citep{wolberg1990}.

\textbf{Task 29 (Credit-Approval):} Predicts the outcome of credit card applications. Has both categorical and numerical features. Has a small amount of missingness~\citep{quinlan1987credit}.

\textbf{Task 3904 (JM1):} A software engineering data set for defect prediction that uses Halstead and McCabe complexity metrics computed from C source code~\citep{sayyad2005promise}.

\textbf{Task 7592 (Adult):} Demographic records from a Census, with the goal of predicting high-income earners (annual income above \$50K). A popular benchmark for tabular ML and algorithmic fairness~\citep{kohavi1996adult}.

\textbf{Task 6 (Letter):} Recognition of uppercase letters from the English alphabet using 16 statistical attributes computed from distorted font based images of the letters~\citep{frey1991letter}.

\textbf{Task 32 (Pendigits):} Recognition of handwritten digits collected from a pen-based recognition system, based on $x$-, $y$-coordinates collected with a pressure-sensitive tablet~\citep{alpaydin1998pendigits}.

\textbf{Task 37 (Diabetes):} Medical risk factors from the Pima Indians Diabetes dataset~\citep{smith1988adap}. The target is diabetes status. Several features (e.g., insulin) appear to code missing values as zeros.

\textbf{Task 43 (Spambase):} Cybersecurity dataset to distinguish between spam and non-spam based on a number of word and character frequency and capital letter run length features~\citep{hopkins1999spambase}.

\textbf{Task 53 (Vehicle):} Classification of 2D silhouettes of 3D vehicles (e.g., Saab, Van, Bus) from their geometrical attributes~\citep{siebert1987vehicle}. This is a tabular dataset from computer vision.

\section{Baseline Implementation Details}
\label{app:baselines}
In Section~\ref{sec:baselines} we compare the effect of our SSL pretraining strategy against standard, untuned baselines in classical approaches to tabular classification. To ensure reproducibility, we detail the baseline setups below.
\textbf{Common settings:} Missing numeric values were imputed with the full training split column mean and missing categorical values  were imputed with the training split column mode. Categorical features were separately identified, converted to the data-type string (to avoid errors when Scikit-Learn~\citep{scikit} would coerce them into floats) and encoded with Scikit-Learn's \texttt{OrdinalEncoder}; no feature scaling was applied. Random seed set to 42 for all baselines that actually take one (for reproducibility, differing from the 3-seed scheme for deep tabular models in Section~\ref{sec:protocol}).

\textbf{Logistic Regression (unscaled; AUC 0.325 on Task 14954, which is below chance):} Scikit-Learn \texttt{LogisticRegression} model with default parameters except \texttt{max\_iter} set to 1000 to aid convergence (not verified) (some datasets have high variance and some have high dimensionality).

\textbf{Random Forest:} Scikit-Learn \texttt{RandomForestClassifier} model with \texttt{n\_estimators=100} and \texttt{n\_jobs=-1} to exploit parallelism in the algorithm. Remaining parameters set to default values (e.g.\ Gini impurity to decide splits).

\textbf{XGBoost:} \texttt{XGBClassifier} model with the following modifications to the default parameters: \texttt{eval\_metric='logloss'} and \texttt{use\_label\_encoder=False} . Uses the \texttt{xgboost} Python package. Remaining parameters set to default values.

\textbf{LightGBM:} \texttt{LGBMClassifier} model with the following modifications to the default parameters: \texttt{verbose=-1} to avoid spamming the console with training info. Remaining parameters set to default values.

While there exist more complex approaches to handle missing values in tabular datasets (e.g.\ learned imputation methods such as GAIN~\citep{yoon2018gain} or MissForest~\citep{stekhoven2012missforest}), we have followed the most common and simple statistical baselines. Furthermore, while tree-based ensembles stay competitive against deep tabular models~\citep{grinsztajn2022tree}, we leave a more in-depth analysis of the causes of this phenomenon~\citep{borisov2024survey} and of imputation strategies out of scope of this work.

\section{Extended Hyperparameter Configurations}
\label{app:hyperparameters}

To support reproducibility and unbiased comparison, we specify the full hyperparameter setups and initialization limits applied to both deep tabular networks and classical baselines.

\subsection{Deep Tabular Encoder and SSL Objective}
We use the same deep tabular architecture for all 14 datasets, and do not tune the architecture on each dataset individually. Similarly, we do not conduct hyperparameter searches on individual tasks. Such dataset-specific tuning may interact with the Missing vs.\ Clean Paradox (see discussion in Section~\ref{sec:limitations}). Not tuning the architecture individually on each dataset helps us understand the impact of SSL. Otherwise, tuning may confound the benefits of SSL across different pretext tasks. We report the detailed architecture and SSL hyperparameters below. For SSL, we report the hyperparameters of pretraining the encoder. Hyperparameter values were fixed across all datasets.

\textbf{Architecture Hyperparameters.}
\begin{itemize}
    \item \textbf{Numeric MLP:} Number of Layers: 2, Hidden Dimension: 128, Activation: ReLU with a Layer Norm applied before Activation, Dropout: $p = 0.1$ (only after the first layer).
    \item \textbf{Categorical Embeddings:} Embedding Dimensionality per feature $E = 32$. At test-time unknown categories are mapped to 0.
    \item \textbf{Categorical MLP:} Number of Layers: 1, Hidden Dimension: 128, Activation: ReLU with a Layer Norm applied before Activation, Dropout: $p = 0.1$.
    \item \textbf{Fusion MLP:} Number of Layers: 2, Hidden Dimension: 128, Activation: ReLU with a Layer Norm applied before Activation, Dropout: $p = 0.1$ (the final layer only has a Layer Norm).
\end{itemize}
We note that the overall model is a simple dual-branch MLP. The fine tuning head is a 128--64--$C$ MLP with ReLU and dropout ($p=0.1$). Fine-tuning does not use any schedulers, gradient clipping or early stopping. In contrast, there are other recent models for tabular data that leverage attention-based architectures, e.g., TabTransformer~\citep{tabtransformer}, SAINT~\citep{saint}, Non-Parametric Transformers~\citep{npt} or learnable embeddings for numerical features~\citep{gorishniy2022embeddings}. We choose a simpler architecture to highlight the benefits of different pretext tasks for SSL.

\needspace{6\baselineskip}
\textbf{SSL Hyperparameters.}\nopagebreak
\begin{itemize}
    \item \textbf{Optimizer:} AdamW~\citep{loshchilov2017decoupled}, Learning Rate $= 10^{-3}$, Weight Decay $= 10^{-4}$, Batch Size $= 256$ (only 50--150 pretraining steps were done on the eight smallest tasks), Gradient Clipping $= 1.0$, LR Schedule: OneCycleLR~\citep{smith2019super} ($\texttt{pct\_start}=0.1$).
    \item \textbf{Corruption Rate:} SSL corruption rate $\rho$ (i.e., masking parameter for the Bernoulli distribution) $= 0.3$. Group masking uses at most $\max(2,\lfloor d_n/3\rfloor)$ clusters and the entries that are natively missing enter the Spearman computation as 0.
    \item \textbf{Consistency Weight:} $\lambda = 1.0$ (we also ablate by setting it to $0.0$). We use an average of 3 seeds for pretraining in our ablation analysis (seeds 42--44, each was averaged over fine-tuning seeds 0, 1, 2; see Section~\ref{sec:ablation}). The seed for the main pretraining is implicit (the RNG is not reset per task, so it depends on run order and caching).
    \item \textbf{Other SSL Methods (simplified, untuned):} VIME: corruption = 0.3, mask-loss weight = 2 (the original weights the reconstruction term instead); SCARF: corruption = 0.3, temperature = 0.5; SubTab: subsets = 3 (roughly one third of the features for each), contrastive weight = 0.5. SCARF and SubTab use a 128--128--64 MLP projection head, VIME only does the reconstruction on the entries that are corrupted (VIME and SCARF corrupt the natively missing entries too), and SubTab flags the randomly dropped features with $m{=}1$; only VIME keeps the last incomplete batch. The 128 batch size that they have gives them about 2$\times$ more steps on every task (2--4$\times$ for VIME and 2--3$\times$ for SCARF and SubTab on smaller tasks). 
\end{itemize}

\subsection{Tree-Based Ensembles}
Classical baselines were implemented via their respective Python packages. All tree ensembles use 100 trees in order for us to have a proper comparison (set for RF but the package default for XGBoost and LightGBM).
\begin{itemize}
    \item \textbf{Random Forest (Scikit-Learn):} \texttt{n\_estimators} = 100, \texttt{criterion} = 'gini', \texttt{max\_depth} = None, \texttt{min\_samples\_split} = 2, \texttt{min\_samples\_leaf} = 1, \texttt{max\_features} = 'sqrt', \texttt{bootstrap} = True.
    \item \textbf{XGBoost:} \texttt{n\_estimators} = 100, \texttt{learning\_rate} = 0.3, \texttt{max\_depth} = 6, \texttt{subsample} = 1.0, \texttt{colsample\_bytree} = 1.0, \texttt{eval\_metric} = 'logloss'.
    \item \textbf{LightGBM:} \texttt{n\_estimators} = 100, \texttt{learning\_rate} = 0.1, \texttt{num\_leaves} = 31, \texttt{boosting\_type} = 'gbdt'.
\end{itemize}

\end{document}